\documentclass[12pt]{article}
\usepackage{amsmath}
\usepackage{graphicx}
\usepackage{natbib}
\usepackage{url} 

\usepackage{placeins}
\usepackage{booktabs}

\usepackage{lineno}

\usepackage[title]{appendix}

\usepackage{pdfpages}
\usepackage[normalem]{ulem}
\usepackage{xcolor}

\usepackage{comment}
\usepackage{sectsty}
\usepackage{lineno}

\usepackage{bm}
\usepackage{amssymb}
\usepackage{dsfont}

\usepackage{authblk}

\newcommand{\blind}{0}

\begin{document}

\def\spacingset#1{\renewcommand{\baselinestretch}%
{#1}\small\normalsize} \spacingset{1}


\if0\blind
{
  \title{\bf Implementing neural network mixed-effects models in Template Model Builder (TMB)}
  \author[1]{Nan Zheng\thanks{Email: k33nz@mun.ca}}
  \author[1]{Hoi Yiu Cheung}
  \author[1]{Vibhu Sharma}
  \author[2]{James T. Thorson}
  \author[3]{Noel G. Cadigan}
  
  \affil[1]{Department of Mathematics and Statistics, Memorial University of Newfoundland, Canada}
  \affil[2]{Resource Ecology and Fisheries Management, Alaska Fisheries Science Center, National Marine Fisheries Service, National Oceanic and Atmospheric Administration, USA}
  \affil[3]{Centre for Fisheries Ecosystems Research, Fisheries and Marine Institute of Memorial University of Newfoundland, Canada}
  \maketitle
} \fi

\if1\blind
{
  \bigskip
  \bigskip
  \bigskip
  \begin{center}
    {\LARGE\bf Implementing neural network mixed-effects models in Template Model Builder (TMB)}
\end{center}
  \medskip
} \fi

\bigskip
\begin{abstract}
	Neural network mixed-effects models (NMMs) have gained traction by combining the strong representation and predictive power of artificial neural networks with the capacity of mixed-effects modeling to capture complex correlation structures. However, existing estimation approaches rely heavily on manual derivations of objective functions and gradients, which inherently forces simplifying approximations and severely constrains the complexity and accuracy of NMMs. In this work, we introduce a general framework for implementing NMMs using Template Model Builder (TMB). By leveraging automatic differentiation and Laplace approximation, TMB requires users to specify only the negative joint log-likelihood and any regularization terms. The framework automatically integrates out random effects and evaluates the marginal objective function alongside its exact gradients, eliminating the need for manual derivations or ad hoc approximations. We demonstrate the efficiency, flexibility, and statistical performance of TMB-based NMMs across two numerical examples, including an application to monotonic NMMs. Reproducible code is provided to facilitate broader adoption.
\end{abstract}

\noindent%
\vfill

\newpage
\spacingset{1.75} 
\section{Introduction}
\label{sec:intro}
The development of technologies for implementing deep neural networks, that is, neural networks with multiple hidden layers, has significantly advanced the fields of artificial intelligence and machine learning \citep{lecun2015deep}. Owing to their strong representational and predictive capabilities, neural networks have been increasingly adopted for analyzing conventional statistical datasets. Compared with modern deep learning applications, these datasets are typically much smaller, making relatively shallow network architectures a practical and commonly adopted choice \citep{hornik1989multilayer,hastie2009elements,goodfellow2016deep}.

A standard feedforward neural network does not explicitly encode structural priors, such as spatial, temporal, or contextual dependencies, within its architecture. As a result, it must learn these relationships solely from the training data, leading to a hypothesis space that is often substantially larger than necessary for structured learning tasks \citep{bengio2009learning,goodfellow2016deep}. Subsequent developments have addressed this limitation by incorporating domain-specific correlation structures into network architectures. For example, convolutional neural networks exploit translation invariance and spatial hierarchies in image data, recurrent neural networks (RNNs) capture temporal dependencies in sequential data, and transformers model long-range contextual relationships in natural language \citep[see, e.g.,][]{chollet2022deep}. Although these architectures substantially improve learning efficiency by embedding appropriate structural assumptions, they typically remain data-intensive and often require training data with specific structural characteristics that are not satisfied in many statistical applications. For example, RNNs are generally designed for regularly sampled sequential data and are most naturally applied to collections of sequences with comparable temporal structure, assumptions that are frequently violated by longitudinal datasets with irregular observation times or unequal sequence lengths \citep{mandel2023neural}. In contrast, statistical mixed-effects models provide considerably greater flexibility for accommodating unbalanced longitudinal data, irregular observation schedules, and complex sampling designs while remaining effective with relatively small sample sizes. Motivated by the complementary strengths of these two paradigms, neural network mixed-effects models (NMMs) have emerged as a natural extension that combines the representational power of neural networks with the flexibility and interpretability of statistical mixed-effects models \citep{tandon2006neural,Tran2017random,mandel2023neural}.

A NMM can be generally formulated as $\bm{\mu}=g\{f_N(\mathbf{X})+\mathbf{Z}^{\top}\mathbf{b}\}$, where $f_N(\mathbf{X})$ represents the output of a neural network with input $\mathbf{X}$, $\mathbf{Z}$ is the design matrix associated with the random effects (REs) $\mathbf{b}$, $\bm{\mu}$ denotes the vector of parameters governing the observational model (e.g., the mean vector), and $g(\cdot)$ is an elementwise transformation analogous to the inverse link function in generalized linear mixed-effects models (GLMMs). Throughout this paper, boldface notation is used for vectors and matrices. The REs $\mathbf{b}$ are typically excluded from the neural network $f_N(\mathbf{X})$, as their inclusion may lead to prohibitive computational costs. Together with the observational model $f(\mathbf{D}|\bm{\mu})$ for responses $\mathbf{D}$, the RE prior $f(\mathbf{b})$, and the weight regularization term $f_{\mathrm{reg}}(\mathbf{w})$, the joint objective function is
\begin{linenomath*} 
	\begin{align}\label{joint_objective_July_27_2026}
		f(\mathbf{D}, \mathbf{b})&=f(\mathbf{D}|\bm{\mu})f(\mathbf{b})f_{\mathrm{reg}}(\mathbf{w}).
	\end{align}
\end{linenomath*} 
The weight regularization term penalizes excessively large neural network weights, thereby reducing overfitting and improving the generalization ability of the model \citep[see, e.g.,][]{krogh1991simple}. Before applying optimization algorithms, such as stochastic gradient descent, in deep learning, the REs $\mathbf{b}$ must be appropriately accommodated to obtain the objective function and its gradients. To this end, \citet{Tran2017random} perform estimation in a Bayesian framework using variational approximation, whereas \citet{mandel2023neural} replace the conditional likelihood function $f(\mathbf{D}|\bm{\mu})$ with a quasi-likelihood function and integrate out $\mathbf{b}$ using the Laplace approximation to obtain the marginal objective function. In this work, we adopt the latter approach because of its computational simplicity and the availability of efficient modeling packages, such as the Template Model Builder \citep[TMB;][]{kristensen2016tmb} within R \citep{Rcoreteam2024language}, which automatically integrates out the REs from the joint likelihood and provides the marginal likelihood together with its gradients. By leveraging automatic differentiation alongside sparse matrix techniques and parallel computing, TMB enables rapid likelihood-based inference even in high-dimensional settings, and offers a highly competitive alternative to Bayesian methods in terms of speed, scalability, and model complexity \citep{thygesen2017validation}. 

Without leveraging modern computational frameworks such as TMB, \citet{mandel2023neural} relied on manual derivations of the marginal objective function and its corresponding gradient to enable numerical optimization. This approach introduces several key limitations. First, as discussed previously, the quasi-likelihood function, rather than the full likelihood, must be used to facilitate manual derivation. Second, the approximation requires omission of a term involving the determinant of the covariance matrix of the REs and the Hessian matrix of the joint objective function with respect to the REs. Neglecting this term may reduce the accuracy and efficiency of inference for certain model parameters. Third, the analytical complexity increases substantially as the number of neural network layers grows, limiting the scalability of the approach \citep[to two layers in][]{mandel2023neural} and making the derivation and implementation increasingly difficult and error-prone. Notably, \cite{mandel2023neural} omitted the derivative of the RE estimators with respect to the model parameters when differentiating the objective function \citep[see, e.g., Eq.~7 in][]{kristensen2016tmb}, an omission that may lead to suboptimal optimization or less precise parameter estimates. 

To address these limitations, we propose and demonstrate the use of TMB for implementing NMMs. Rather than using a quasi-likelihood approximation, we directly provide the joint objective function (\ref{joint_objective_July_27_2026}) to TMB. TMB then automatically and efficiently computes the Laplace approximation to the marginal objective function, together with its corresponding gradients, without omitting any terms. Consequently, the proposed approach improves computational efficiency, inferential accuracy, ease of implementation, and scalability relative to the manual derivation approach.

The remainder of this paper is organized as follows. Section \ref{sec:meth} details the proposed NMM framework and its implementation using TMB. Section \ref{sec:simulation_August_4_21_2026} replicates the simulation studies from \cite{mandel2023neural}, demonstrating the computational efficiency and superior accuracy of our approach. In Section \ref{sec:real_data_3LNO_AP_maturity}, we apply both the NMM and GLMM to an American plaice maturity dataset, illustrating the ease of implementing monotonic neural networks within this framework. Finally, Section \ref{sec:conc} offers concluding remarks and directions for future research.

\section{Methods}
\label{sec:meth}
\begin{figure}[htbp]
	\centering
	\includegraphics[width=\textwidth]{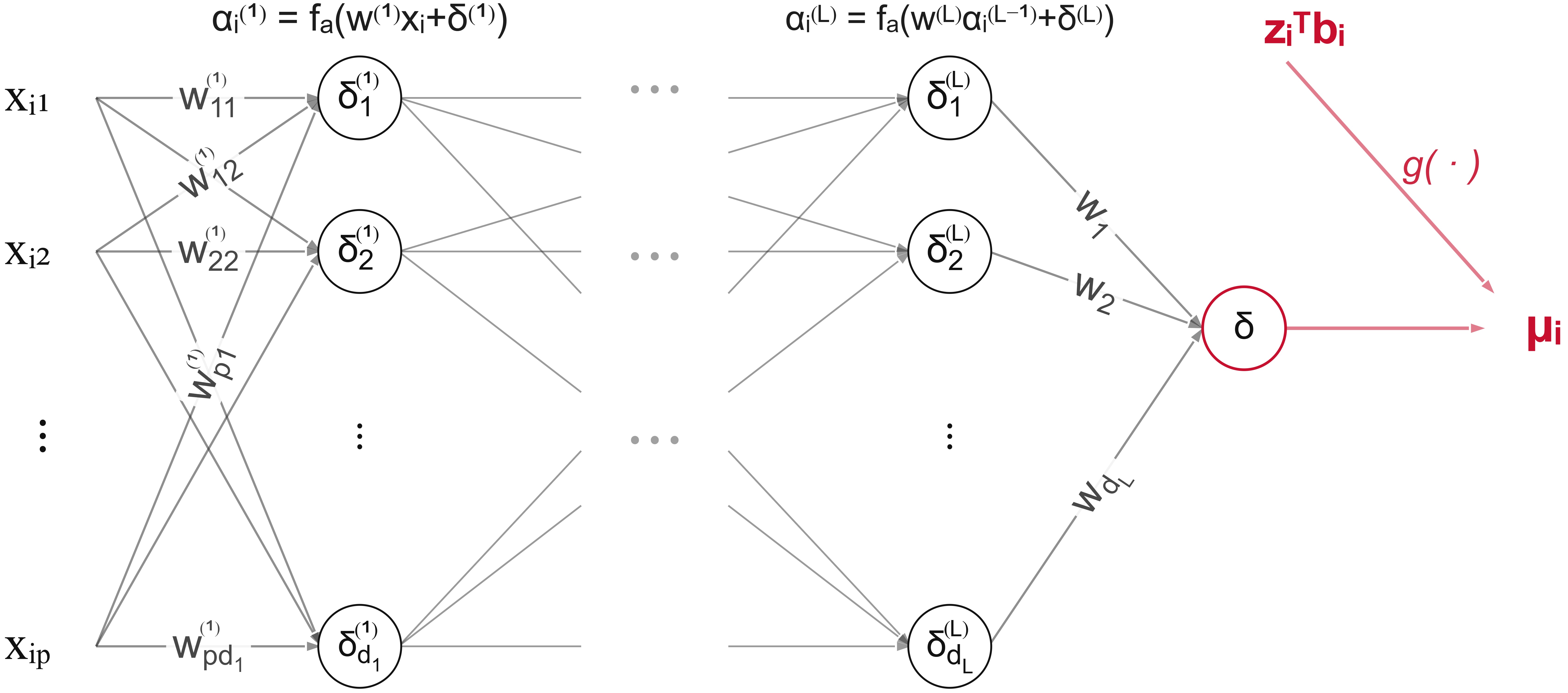}
	\caption{The architecture of a multi-layer NMM.}
	\label{fig:NMM_architecture_August_18_2026}
\end{figure}
\noindent Our model architecture is displayed in Fig. \ref{fig:NMM_architecture_August_18_2026} and closely follows the feed-forward network framework of \citet{mandel2023neural}. Let $\mathbf{x}_i \in \mathbb{R}^p$ denote the input vector for the $i$th observation and $\mu_i \in \mathbb{R}$ be its corresponding univariate output. For a network with $L$ hidden layers, where layer $l \in \{1, \dots, L\}$ consists of $d_l$ units, each layer sequentially applies an affine transformation followed by an element-wise activation function $f_a(\cdot)$ \citep[see, e.g.,][]{chollet2022deep}. Specifically, for the $i$th observation, the output of the first hidden layer, $\bm{\alpha}^{(1)}_i \in \mathbb{R}^{d_1}$, is expressed as
\begin{linenomath*}
	\begin{align}\label{layer_1_output_July_29_2026}
		\bm{\alpha}^{(1)}_{i} &= f_a\{\mathbf{w}^{(1)}\mathbf{x}_i + \bm{\delta}^{(1)}\},
	\end{align}
\end{linenomath*}
where $\mathbf{w}^{(1)} \in \mathbb{R}^{d_1 \times p}$ and $\bm{\delta}^{(1)} \in \mathbb{R}^{d_1}$ represent the weight matrix and bias vector, respectively. For each hidden layer $l \in \{2, \dots, L\}$, the activation output is given by
\begin{linenomath*}
	\begin{align}\label{layer_l_output_July_29_2026}
		\bm{\alpha}^{(l)}_{i} &= f_a\{\mathbf{w}^{(l)}\bm{\alpha}^{(l-1)}_{i} + \bm{\delta}^{(l)}\},
	\end{align}
\end{linenomath*}
with parameters $\mathbf{w}^{(l)} \in \mathbb{R}^{d_l \times d_{l-1}}$ and $\bm{\delta}^{(l)} \in \mathbb{R}^{d_l}$. Finally, the univariate output of the neural network is related to the observational parameter $\mu_i$ via
\begin{linenomath*}
	\begin{align}\label{mui_July_30_2026}
		\mu_i(\mathbf{b}_i) &= g\{\mathbf{w}\,\bm{\alpha}^{(L)}_{i} + \delta + \mathbf{Z}_i^{\top}\mathbf{b}_i\},
	\end{align}
\end{linenomath*}
where $\mathbf{w} \in \mathbb{R}^{1 \times d_L}$, $\delta \in \mathbb{R}$, and $\mathbf{Z}_i$ and $\mathbf{b}_i$ denote the design matrix and RE vector associated with the $i$th observation, respectively. The REs $\mathbf{b}_i$ are assumed to follow a multivariate normal (MVN) distribution, inducing dependence among the components of the $i$th observation. Dependence across observations can be introduced either by sharing one or more RE components among observations or by specifying a correlated MVN distribution for the REs $\mathbf{b}_i$ across observations. In \citet{mandel2023neural}, $\mu_i$ denotes the conditional mean of the response under an exponential-family distribution given $\mathbf{b}_i$, which enables the use of a quasi-likelihood approximation to the true conditional likelihood. In contrast, our framework is not restricted to exponential-family distributions. The observational distribution can be specified more generally, and $\mu_i$ need not represent the conditional mean; rather, it may denote any governing parameter of the observational distribution. 

A comparison can also be made to the GLMM framework \citep{breslow1993generalized}, in which the conditional mean is linked to the linear predictor via\begin{linenomath*}\begin{align}\label{GLMM_link_July_30_2026}\mu_i &= g\{\mathbf{x}^{\top}_i\bm{\beta} +\beta_0+ \mathbf{Z}_i^{\top}\mathbf{b}_i\},\end{align}\end{linenomath*}
where $\bm{\beta}$ represents the vector of fixed effects, $\mu_i$ denotes the conditional mean of the $i$th observation given the REs, and $g(\cdot)$ is the inverse link function. The NMM defined in Eqs.~(\ref{layer_1_output_July_29_2026})--(\ref{mui_July_30_2026}) replaces the fixed-effects portion of the linear predictor with the output of a feed-forward neural network. Because neural networks offer significantly greater expressiveness than linear predictors, the NMM is expected to yield better representational and predictive capabilities compared to the standard GLMM.

Let there be $n$ observations, and let $\mathbf{B}=(\mathbf{b}_1^{\top},\ldots,\mathbf{b}_n^{\top})^{\top}$ denote the collection of all REs. The marginal likelihood $l$ for NMM defined by (\ref{layer_1_output_July_29_2026})--(\ref{mui_July_30_2026}) is expressed as
\begin{linenomath*} 
	\begin{align}\label{marginal_likelihood_July_31_2026}
	l(\bm{\theta}|\mathbf{D}_1,\ldots, \mathbf{D}_n) &=	\int\left[ \prod_{i=1}^n f\{\mathbf{D}_i|\mu_i(\mathbf{b}_i)\}\right] f(\mathbf{B}) d\mathbf{B},
	\end{align}
\end{linenomath*} 
where $\mathbf{D}_i$ denotes the $i$th observation, $f\{\mathbf{D}_i\mid\mu_i(\mathbf{b}_i)\}$ is its conditional distribution given the REs, and $f(\mathbf{B})$ is the prior (marginal) distribution of $\mathbf{B}$. The vector $\bm{\theta}$ collects all model parameters, including the neural network weights and biases, observational model parameters, and prior distribution parameters. To evaluate the integral over the REs in Eq.~(\ref{marginal_likelihood_July_31_2026}), we employ TMB. TMB efficiently approximates the marginal likelihood via the Laplace approximation, scaling to tens of thousands of REs by exploiting the sparsity in their joint distribution with the observations.

The model parameters are estimated by minimizing the objective function
\begin{linenomath*} 
	\begin{align}\label{loss_August_2_2026}
		 &-\log\{l(\bm{\theta}|\mathbf{D}_1,\ldots, \mathbf{D}_n)\} + \lambda(\mathbf{W}^{\top}\mathbf{W} + \bm{\Delta}^{\top}\bm{\Delta}),
	\end{align}
\end{linenomath*} 
where $\mathbf{W}$ and $\bm{\Delta}$ denote the vectors of all neural network weights and biases, respectively, and $\lambda > 0$ is a regularization hyperparameter. The second term introduces $L_2$ regularization, which penalizes larger parameter magnitudes to mitigate overfitting and enhance generalization. Following \cite{mandel2023neural}, we employ an $L_2$ penalty to prioritize predictive accuracy over feature selection \citep{tibshirani1996regression, goodfellow2016deep} while ensuring differentiability.

\section{Simulation study}
\label{sec:simulation_August_4_21_2026}
We assess the proposed TMB implementation of the NMM by reproducing the nonlinear simulation experiment of \citet{mandel2023neural}, considering both a continuous and a binary response. In each case the network is provided only with the raw predictors and must recover a nonlinear signal from them, and predictive performance is assessed by leave-one-out cross-validation on the final observation. To facilitate direct comparison with the results of \citet{mandel2023neural}, we adopt the notation used in their simulation studies throughout this section.

\subsection{Simulation design}
\label{subsec:sim_design_August_5_2026}
For each of $m=100$ subjects, the number of observations $n_i$ is drawn from a $\mathrm{Poisson}(6)+2$ distribution, ensuring at least one training and one validation observation per subject. For each observation we generate six independent binary predictors $\mathbf{x}_{ij}=(x_{ij1},\ldots,x_{ij6})^{\top}$, each from a $\mathrm{Bernoulli}(0.5)$ distribution, and form three nonlinear features
\begin{linenomath*}
	\begin{align}\label{XOR_features_August_5_2026}
		Z_{ij1}=\mathds{1}(x_{ij1}+x_{ij2}=1),\quad
		Z_{ij2}=\mathds{1}(x_{ij3}+x_{ij4}\neq 1),\quad
		Z_{ij3}=\mathds{1}(x_{ij5}+x_{ij6}=1),
	\end{align}
\end{linenomath*}
which encode exclusive-or (XOR) type interactions that a linear predictor cannot represent. Writing $\mathbf{Z}_{ij}=(Z_{ij1},Z_{ij2},Z_{ij3})^{\top}$, the continuous response is generated as
\begin{linenomath*}
	\begin{align}\label{cont_datagen_August_5_2026}
		y_{ij}=\mathbf{Z}_{ij}^{\top}\bm{\beta}+b_i+\epsilon_{ij},\qquad \bm{\beta}\sim\mathrm{MVN}(2\cdot\mathbf{1},\mathbf{I}),
	\end{align}
\end{linenomath*}
and the binary response as
\begin{linenomath*}
	\begin{align}\label{bin_datagen_August_5_2026}
		\mathrm{logit}\{\mathrm{P}(y_{ij}=1)\}=\mathbf{Z}_{ij}^{\top}\bm{\beta}+b_i+\epsilon_{ij},\qquad \bm{\beta}\sim\mathrm{MVN}(3\cdot\mathbf{1},\mathbf{I}),
	\end{align}
\end{linenomath*}
where the subject-level random intercept is $b_i\sim N(0,\tau)$ and $\epsilon_{ij}\sim N(0,0.05^2)$, and $\bm{\beta}$ is drawn once per replicate. The variance component $\tau$ governs the strength of within-subject correlation and is swept over $\{0,0.1,0.2,0.3,0.4,0.5\}$ for the continuous response and $\{0,4,8,12,16,20\}$ for the binary response, following \citet{mandel2023neural}. Crucially, the model is fit using only the raw predictors $\mathbf{x}_{ij}$; the features $\mathbf{Z}_{ij}$ are never supplied, so the network must recover the XOR structure of \eqref{XOR_features_August_5_2026} on its own.

Predictive accuracy is evaluated by leave-one-out cross-validation: the final observation of each subject is withheld, the model is fit to the remaining observations, and the withheld observation is predicted as $\hat{\mu}_{ij}=g\{\hat{f}_N(\mathbf{x}_{ij})+\hat{b}_i\}$, combining the fitted network output with the subject's estimated random intercept. Performance is summarized by the mean squared prediction error (MSPE) for the continuous response and by the area under the receiver operating characteristic curve (AUC) for the binary response, averaged across the 100 withheld observations. Each configuration is replicated 500 times.

\subsection{Model and estimation}
\label{subsec:sim_model_August_5_2026}
We implement the two-hidden-layer network of \citet{mandel2023neural}, denoted GNMM-2, with $d_1=10$ units in the first hidden layer and $d_2=5$ in the second, a six-dimensional input, and a scalar output, as in Eqs.~(\ref{layer_1_output_July_29_2026})--(\ref{mui_July_30_2026}). The activation function $f_a$ is the logistic sigmoid, and the output transformation $g$ is the identity for the continuous response and the inverse logit for the binary response. A single scalar random intercept enters the output layer for each subject ($\mathbf{Z}_i\equiv 1$ in Eq.~(\ref{mui_July_30_2026})), with prior $b_i\sim N(0,\tau)$. The observational distribution is Gaussian with an estimated residual variance for the continuous response and Bernoulli for the binary response, as specified in Eqs. (\ref{cont_datagen_August_5_2026}) and (\ref{bin_datagen_August_5_2026}), respectively.

Estimation follows the objective (\ref{loss_August_2_2026}) with regularization parameter $\lambda=0.005$, matching the value used for the two-hidden-layer network in \citet{mandel2023neural}. The random intercepts are integrated out by TMB via the Laplace approximation, and the resulting marginal objective is optimized using \texttt{nlminb} function in R; standard deviations are parameterized on the log scale to enforce positivity, and $\lambda$ is held fixed. In contrast to the manual quasi-likelihood derivation of \citet{mandel2023neural}, the TMB implementation requires only a specification of the joint objective (\ref{joint_objective_July_27_2026}), automatically deriving the marginal objective and its gradients.

\subsection{Results}
\label{subsec:sim_results_August_5_2026}
\begin{figure}[htbp]
	\centering
	\includegraphics[width=\textwidth]{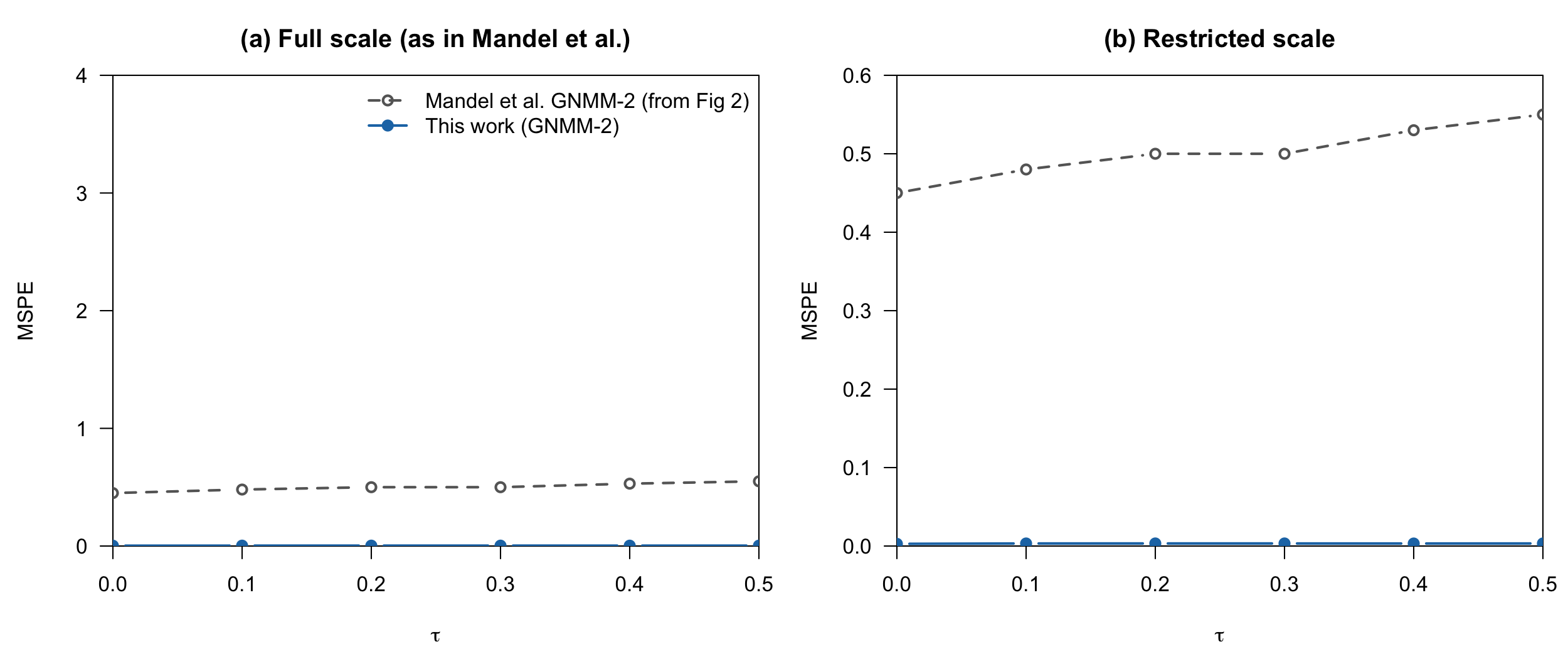}
	\caption{Mean squared prediction error (MSPE) of the TMB implementation of GNMM-2 for the continuous response (solid, blue), as a function of the random-intercept variance $\tau$, overlaid on the GNMM-2 results reported by \citet{mandel2023neural} (grey dashed, read from their Figure~2). Both panels show the same two curves for the nonlinear data-generating mechanism and a first hidden layer of 10 units. Panel~(a) uses the full $y$-axis range $[0,4]$ of \citet{mandel2023neural}: on this scale the TMB curve is pinned to the horizontal axis, clearly far below the reported GNMM-2 curve. Panel~(b) shows the same data on a restricted $y$-axis $[0,0.6]$, which makes the difference explicit: the TMB MSPE (${\approx}\,0.003$) lies just above the axis, well below the reported GNMM-2 MSPE (${\approx}\,0.45$--$0.55$).}
	\label{fig:sim_mspe}
\end{figure}

\begin{figure}[htbp]
	\centering
	\includegraphics[width=0.68\textwidth]{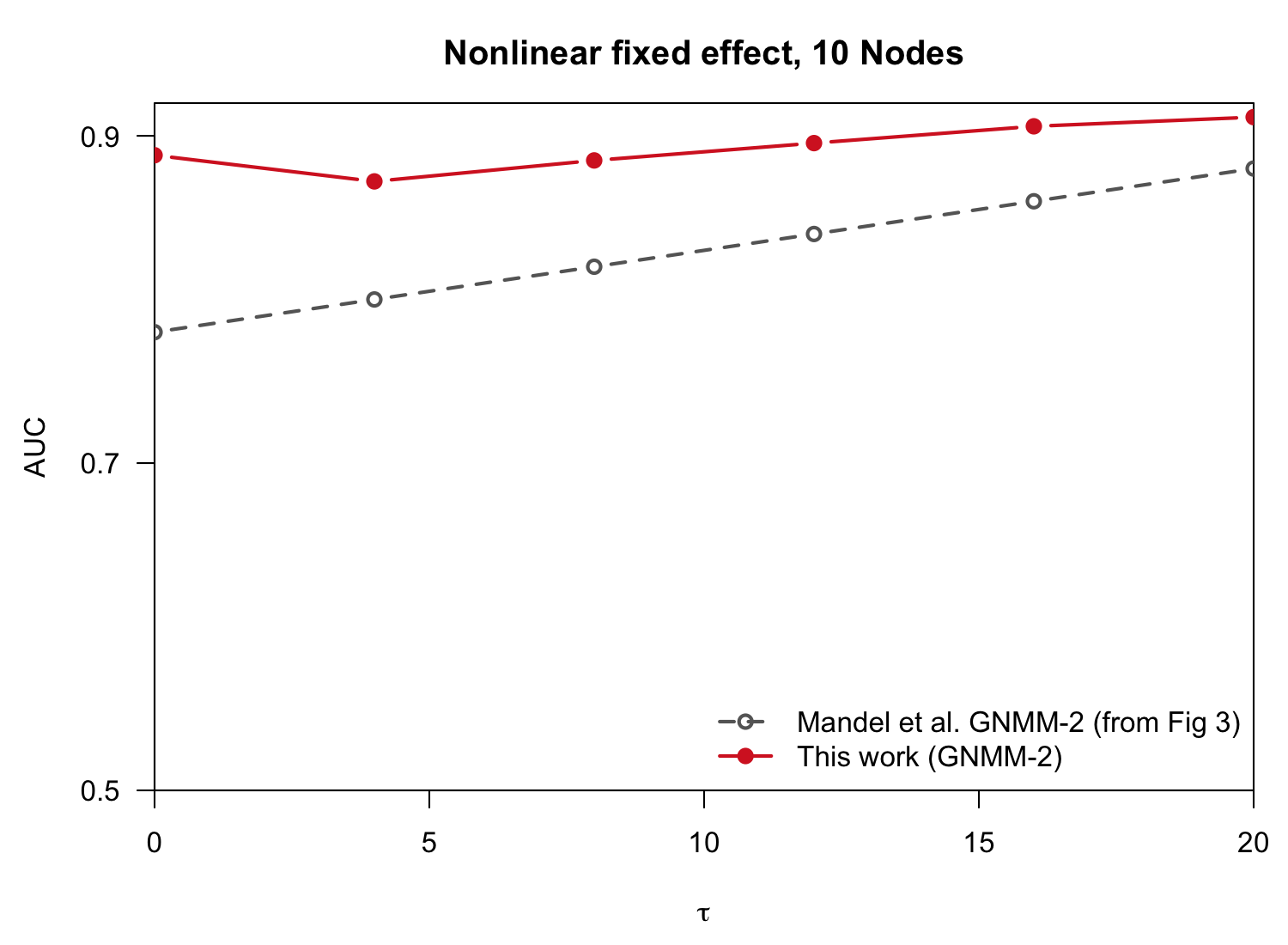}
	\caption{Area under the ROC curve (AUC) of the TMB implementation of GNMM-2 for the binary response (solid), as a function of the random-intercept variance $\tau$, overlaid on the GNMM-2 results reported by \citet{mandel2023neural} (grey dashed, read from their Figure~3). Both correspond to the nonlinear data-generating mechanism and a first hidden layer of 10 units; the axes match the scale of \citet{mandel2023neural}.}
	\label{fig:sim_auc}
\end{figure}

Figures~\ref{fig:sim_mspe} and~\ref{fig:sim_auc} present the predictive performance of the TMB implementation of GNMM-2 as a function of $\tau$, overlaid on the corresponding GNMM-2 results reported by \citet{mandel2023neural}. For the continuous response, the TMB MSPE is essentially flat at approximately $0.003$ across all values of $\tau$, closely tracking the irreducible noise floor $\sigma_\epsilon^2=0.0025$: the fitted network recovers the XOR signal almost exactly and the random intercept absorbs the within-subject correlation, leaving only observation noise. For the binary response, the TMB AUC rises from approximately $0.89$ at $\tau=0$ to $0.91$ at $\tau=20$, increasing with $\tau$ because a larger random-intercept variance renders the subject-specific estimate $\hat{b}_i$ more informative for the withheld observation. 

In both settings the TMB implementation attains markedly better predictive performance than the GNMM-2 results reported by \citet{mandel2023neural}, whose MSPE is approximately $0.5$ and whose AUC ranges from approximately $0.78$ to $0.88$. 

\FloatBarrier

\section{Real data example: maturity data for 3LNO American plaice}\label{sec:real_data_3LNO_AP_maturity}
To illustrate the implementation of the NMM using real data, we fitted the model to maturity data for female American plaice (\textit{Hippoglossoides platessoides}) collected during annual surveys conducted by Fisheries and Oceans Canada (DFO) in the Northwest Atlantic Fisheries Organization (NAFO) Divisions 3L, 3N, and 3O (3LNO) from 1978 to 2015. Fish born in the same year (i.e., belonging to the same cohort) are generally exposed to similar environmental conditions and other shared influences throughout their life history. Therefore, cohort (birth year), rather than survey year, was used as the temporal index. In total, the dataset comprises 58 cohorts spanning birth years from 1958 to 2015. We further partition the 58 cohorts into cohort segments (cohort\_seg) of eight consecutive cohorts each. Specifically, the first eight cohorts are assigned cohort\_seg 1, the next eight cohorts are assigned cohort\_seg 2, and so forth. Under this partitioning, the final cohort segment contains only three cohorts. To avoid such a small segment, we merge it with the penultimate cohort segment (cohort\_seg 7). Consequently, the data are divided into seven cohort segments, with the first six segments each containing eight consecutive cohorts and the seventh segment containing 11 consecutive cohorts.

The NMM consisted of two hidden layers, each containing twenty nodes. Input for the $i$th fish was an eight-dimensional vector comprising age as the first component, followed by a seven-dimensional one-hot encoded representation of the cohort segment (cohort\_seg number), where the active category was assigned a value of 1 and all remaining categories were set to 0. The activation function $f_a$ in Fig.~\ref{fig:NMM_architecture_August_18_2026} is the logistic sigmoid. The output yields the conditional maturation probability for the $i$th fish given the REs, specifying Eq. (\ref{mui_July_30_2026}) as
\begin{linenomath*}
	\begin{align}\label{NMM_AP_August_5_2026}
		\mathrm{P}_i(a_i|\bm{\theta},b_{c_i}) &= \mathrm{inv\_logit}\{\mathbf{w}\,\bm{\alpha}^{(2)}_{i} + \delta + b_{c_i}\},
	\end{align}
\end{linenomath*}
where $a_i$ and $c_i$ denote respectively the age and cohort index for the $i$th individual, $\mathrm{inv\_logit}(\cdot)$ represents the inverse logit transformation function, and cohort-specific REs ($b_1, \dots, b_{58}$) are modeled using a first-order autoregressive [AR(1)] process across the 58 cohorts. Maturity was recorded as a binary response variable ($d_i$), where 1 indicates a mature fish and 0 indicates an immature fish. Accordingly, the observational model $f\{\mathbf{D}_i\vert{}\mu_i(\mathbf{b}_i)\}$ in Eq. (\ref{marginal_likelihood_July_31_2026}) takes the Bernoulli form: $f\{d_i|\mathrm{P}_i\} = \{\mathrm{P}_i\}^{d_i}\{1-\mathrm{P}_i\}^{1-d_i}$. The objective function is defined by Eqs. (\ref{marginal_likelihood_July_31_2026})-(\ref{loss_August_2_2026}).

The maturation probability in Eq. (\ref{NMM_AP_August_5_2026}) is constrained to be a non-decreasing function of age. This monotonicity constraint is imposed by generalizing the Lagrange multiplier method using the Karush-Kuhn-Tucker (KKT) conditions \citep[see, e.g.,][]{karush1939minima,chen2022monotonic}. Specifically, let
\begin{linenomath*}
	\begin{align}\label{max_derivative_August_6_2026}
		h(\bm{\theta}) &= \sup_{a,c} \left\lbrace -\dfrac{\partial \mathrm{P}(a|\bm{\theta},b_{c})}{\partial a}\right\rbrace ,
	\end{align}
\end{linenomath*}
denote the minimum derivative of the maturity probability with respect to age $a$ across all the ages and cohort segments ($c$). The generalized Lagrangian function is then formulated as
\begin{linenomath*}
	\begin{align}\label{general_Lagrange_fn_August_6_2026}
		\mathcal{L}(\bm{\theta},\eta) &= -\log\{l(\bm{\theta}|d_1,\ldots, d_n)\} + \lambda(\mathbf{W}^{\top}\mathbf{W} + \bm{\Delta}^{\top}\bm{\Delta}) + \eta h(\bm{\theta}),
	\end{align}
\end{linenomath*}
where $\eta \ge 0$ is the KKT multiplier corresponding to the inequality constraint. According to the KKT conditions, the optimization procedure begins by setting $\eta = 0$ to find the initial minimizer $\hat{\bm{\theta}}$ of the generalized Lagrangian function (\ref{general_Lagrange_fn_August_6_2026}) with respect to $\bm{\theta}$. The algorithm then evaluates whether the monotonicity constraint $h(\hat{\bm{\theta}}) \le 0$ is satisfied. If the condition is violated, $\eta$ is increased by a small increment and $\hat{\bm{\theta}}$ is re-estimated; this process is repeated iteratively until the condition $h(\hat{\bm{\theta}}) \le 0$ holds. TMB efficiently computes the generalized Lagrangian function (\ref{general_Lagrange_fn_August_6_2026}) and its gradient with respect to $\bm{\theta}$ for optimization via \texttt{nlminb} function in R. 

As a baseline for comparison, we consider a GLMM, in which the conditional maturation probability for the $i$th fish, given the REs, is modeled as:
\begin{linenomath*}
	\begin{align}\label{GLMM_AP_August_12_2026}
		\mathrm{P}_i(a_i|\beta_0, \beta_{s_{c_i}},b_{c_i}) &= \mathrm{inv\_logit}(\beta_0 + \beta_{s_{c_i}}a_i + b_{c_i}),
	\end{align}
\end{linenomath*}
where $s_{c_i}$ is the cohort\_seg number for cohort $c_i$, and $\beta_0$ and $\beta_{s_{c_i}}$'s are model parameters to estimate. The principal distinction between the NMM (\ref{NMM_AP_August_5_2026}) and the GLMM (\ref{GLMM_AP_August_12_2026}) lies in the age effect: while the latter incorporates a strictly linear predictor with respect to age, the former accommodates flexible, nonlinear functional relationships.

We used the 3L data (31,427 observations) as the training set, the 3N data (23,392 observations) as the validation set, and the 3O data (24,104 observations) as the test set. We evaluated model performance on the validation and test sets using the average negative log-likelihood, referred to as the binary cross-entropy (BCE) in the machine learning literature. For instance, the BCE for the validation set, denoted by $\mathrm{BCE}_{\mathrm{valid}}$, is defined as
\begin{linenomath*}
	\begin{align}\label{BS_August_16_2026}
		\text{BCE}_{\mathrm{valid}} &= -\dfrac{1}{N_{\mathrm{valid}}} \sum_{i=1}^{N_{\mathrm{valid}}} \left[ d_i \log \{\mathrm{P}i(a_i \mid \hat{\bm{\varphi}}, \hat{b}_{c_i})\} + (1-d_i)\log \{1-\mathrm{P}i(a_i \mid \hat{\bm{\varphi}}, \hat{b}_{c_i})\} \right],
	\end{align}
\end{linenomath*}
where $i = 1, \dots, N_{\mathrm{valid}}$ indexes the validation observations, and $\hat{\bm{\varphi}}$ and $\hat{b}_{c_i}$ represent the estimated model parameters and REs, respectively. 

We implemented both the NMM and GLMM using TMB, which efficiently evaluates the marginal objective function and its gradient. These quantities were then supplied to \texttt{nlminb()} for optimization of the marginal objective function. During optimization, the models were fitted to the training data, while $\text{BCE}_{\mathrm{valid}}$ was recorded at each optimization step. We identified the point at which $\text{BCE}_{\mathrm{valid}}$ ceased to improve as the point of best generalization. The corresponding parameter and RE estimates were subsequently taken as the final estimates. Prior to this final fit, we performed a grid search over the regularization hyperparameter $\lambda$, and at each grid point, the KKT multiplier $\eta$ was determined using the algorithm described previously. This procedure yielded $\lambda=0.001$ and $\eta=1.7$. 

Based on the optimized parameter and RE estimates, we evaluated the BCE on the test set, $\text{BCE}_{\mathrm{test}}$, to compare the predictive performance of the NMM and GLMM. The NMM achieved a slightly lower $\text{BCE}_{\mathrm{test}}$ of $0.249$, compared with $0.250$ for the GLMM.

\begin{figure}[htbp]
	\centering
	\includegraphics[width=\textwidth]{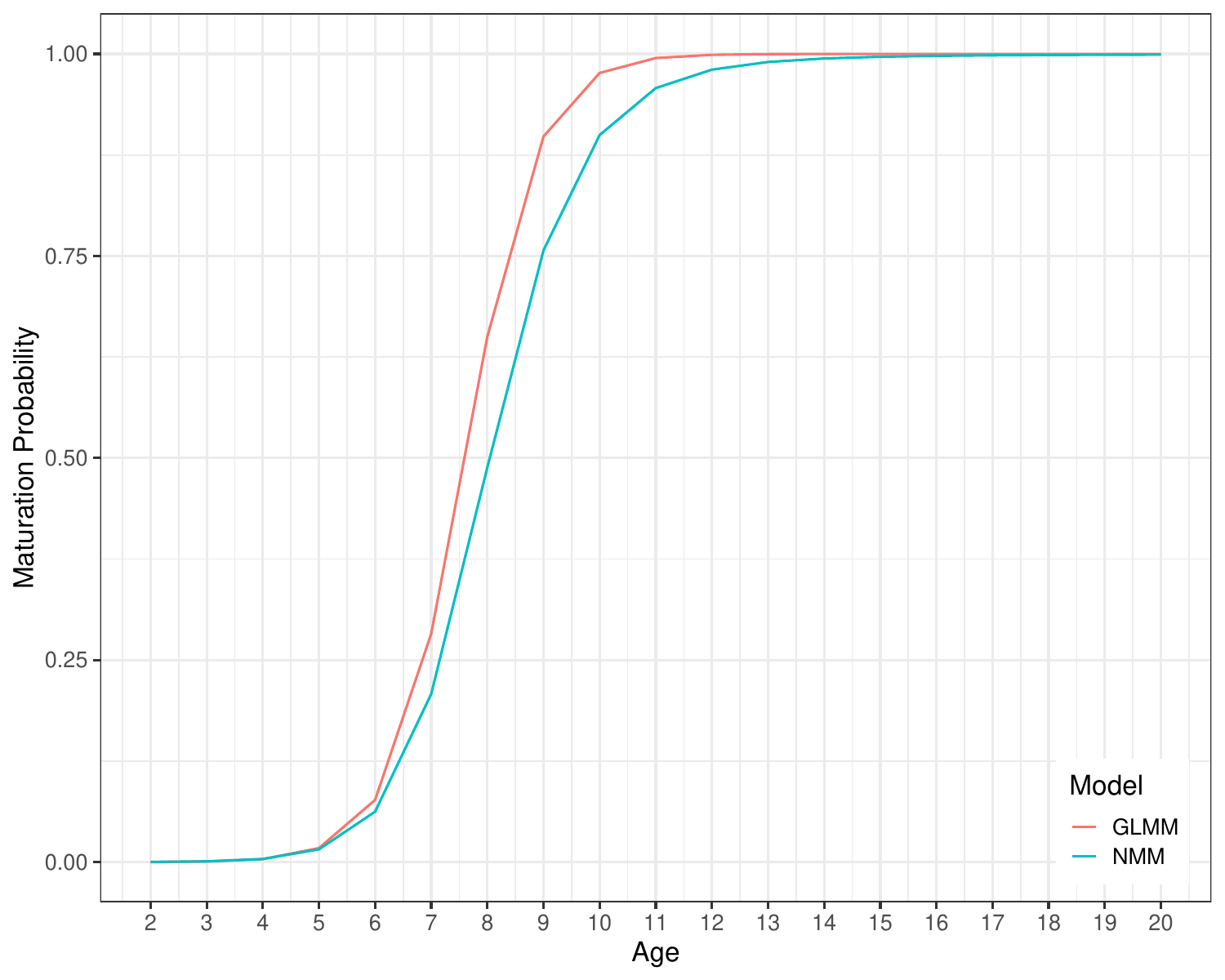}
	\caption{Predicted probabilities of maturation by age for the 50th cohort, using the NMM and GLMM.}
	\label{fig:NMM_GLMM_comparison_cohort_50_August_18_2026}
\end{figure}

TMB automatically predicts the REs using their posterior modes and predicts functions of the REs and fixed parameters using plug-in predictors. Figure \ref{fig:NMM_GLMM_comparison_cohort_50_August_18_2026} compares the predicted age-specific maturation probabilities for the 50th cohort between the NMM and the GLMM. While the GLMM yields predictions that are symmetric around the inflection point, the NMM accommodates mild asymmetry to better capture the underlying data.
 
\FloatBarrier
\section{Conclusion}
\label{sec:conc}
Due to its enormous implementation and computation complexity, efficient software and hardware tools are always one of the central problem in artificial intelligence. In this work, we propose to implement neural network mixed-effects models (NMMs) in statistical modeling using Template Model Builder (TMB). In this approach, users need to specify only the negative joint log-likelihood and any regularization terms. The framework automatically integrates out random effects and evaluates the marginal objective function alongside its exact gradients, eliminating the need for manual derivations or ad hoc approximations.

Replicating the simulation studies of \citet{mandel2023neural}, the results in Section~\ref{sec:simulation_August_4_21_2026} highlight several advantages of our proposed methodology. First, adopting a full likelihood formulation instead of a quasi-likelihood approach increases statistical efficiency. Second, by fully automating the derivation process, our method eliminates human error and avoids the heuristic omissions or approximations typically needed for manual solutions. As a result, our approach offers enhanced estimation accuracy, improved implementation efficiency, and superior scalability toward complex NMMs.

The empirical results in Section~\ref{sec:real_data_3LNO_AP_maturity} demonstrate the simplicity of implementing constrained NMMs within the TMB framework. Specifically, TMB enables the efficient computation of the monotonicity-constrained objective function and its gradients for the downstream optimization algorithm. Furthermore, even without extensive architecture tuning, this application underscores the representational capacity of NMMs in modeling intricate, nonlinear input–output dynamics.

In the current implementation, feedforward neural networks are coded manually; however, extending this manual approach to convolutional, recurrent, or transformer architectures is impractical. Immediate future work will focus on integrating Keras \citep[see, e.g., Chapter 3 of][]{chollet2022deep}, an R interface to TensorFlow, with RTMB \citep{RTMB}, which provides native R automatic differentiation for TMB without C++ compilation. Synthesizing these tools will allow us to construct sophisticated deep learning mixed-effects models. A secondary direction involves extending these models to incorporate complex spatiotemporal structures via random effects. Additionally, migrating our implementation of NMMs from a strictly CPU-bound environment to a hybrid CPU/GPU workflow will optimize computational performance and support the increased overhead of these advanced architectures. 

\FloatBarrier
\section*{Codes availability}
The TMB code supporting the findings of this study is publicly available at\\ https://github.com/ZhengNan18/Implementing-neural-network-mixed-effects-models-in-Template-Model-Builder-TMB-
\section*{Acknowledgments}
Research funding to NZ was provided by the Natural Sciences and Engineering Research Council of Canada [RGPIN-2024-04746].
Research funding to NC was provided by 1) the Ocean Choice
International Industry Research Chair program at the Marine Institute
of Memorial University of Newfoundland, 2) the Ocean Frontier Institute, through an award from the Canada First Research Excellence Fund, and 3) the Natural Sciences and Engineering Research Council of Canada [RGPIN-2016-04307].

\bibliographystyle{chicago}

\bibliography{mybib_marine}


\end{document}